\pdfoutput=1
\documentclass[11pt]{article}

\usepackage[]{acl}

\usepackage{times}
\usepackage{latexsym}

\usepackage[T1]{fontenc}
\usepackage[utf8]{inputenc}

\usepackage{microtype}

\usepackage{times}
\usepackage{soul}
\usepackage{url}
\usepackage[utf8]{inputenc}
\usepackage{graphicx}
\usepackage{amsmath}
\usepackage{amsthm}
\usepackage{booktabs}
\usepackage{algorithm}
\usepackage{algorithmic}
\usepackage{multirow}
\usepackage{subfigure}
\usepackage{natbib}
\usepackage{amsmath}
\usepackage{diagbox}
\usepackage{color}
\usepackage[switch]{lineno}
\graphicspath{{figure/}}

\title{IMCI: Integrate Multi-view Contextual Information \\ for Fact Extraction and Verification}

\author{Hao Wang$^{\dag}$, Yangguang Li$^{\ddag}$, Zhen Huang$^{\dag}$, Yong Dou$^{\dag}$ \\
  $^{\dag}$  National University of Defense Technology, Changsha \\
  $^{\ddag}$ SenseTime, Beijing\\
  \texttt{\{hao.wang, zhenhuang, yongdou\}@nudt.edu.cn}\\
  \texttt{liyangguang@sensetime.com}}

\begin{document}
\maketitle

\begin{abstract}
With the rapid development of automatic fake news detection technology, fact extraction and verification (FEVER) has been attracting more attention. 
The task aims to extract the most related fact evidences from millions of open-domain Wikipedia documents and then verify the credibility of corresponding claims.
Although several strong models have been proposed for the task and they have made great progress, we argue that they fail to utilize multi-view contextual information and thus cannot obtain better performance.
In this paper, we propose to integrate multi-view contextual information (IMCI) for fact extraction and verification.
For each evidence sentence, we define two kinds of context, i.e. \textbf{intra-document context} and \textbf{inter-document context}.
Intra-document context consists of the document title and all the other sentences from the same document.
Inter-document context consists of all other evidences which may come from different documents.
Then we integrate the multi-view contextual information to encode the evidence sentences to handle the task.
Our experimental results on FEVER 1.0 shared task show that our IMCI framework makes great progress on both fact extraction and verification, and achieves state-of-the-art performance with a winning FEVER score of $72.97\%$ and label accuracy of $75.84\%$ on the online blind test set.
We also conduct ablation study to detect the impact of multi-view contextual information.
Our codes will be released at \url{https://github.com/phoenixsecularbird/IMCI}.
\end{abstract}

\section{Introduction}

\begin{figure}[!h]
\centerline{\includegraphics[scale=0.53]{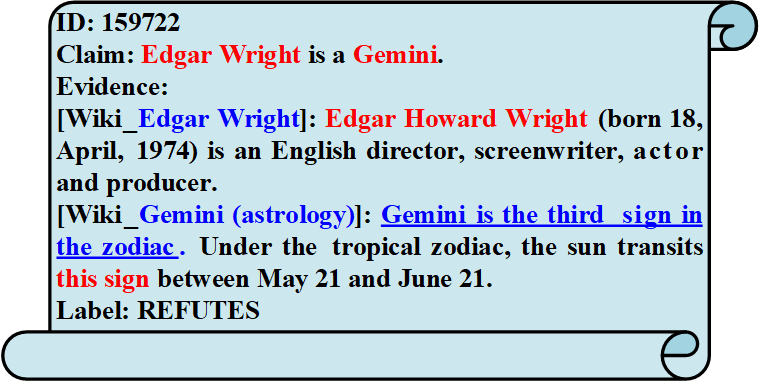}}
\caption{An example from FEVER 1.0 shared task. (\textbf{Underlined sentence} is the \textbf{unlabeled intra-document context}. Words in red involve alias name, coreference and multi-hop reasoning, which may lead to model confusion. Words in blue help to handle these issues.)}
\label{examples}
\end{figure}

Fake news propagation is a severe social problem, which may cause great loss and lead to serious consequence, e.g. panic, quarrel, opposition and even war.
The situation has become a general concern since Brexit and the U.S. President Campaign in 2016 and gets far more intense due to COVID-19 pandemic \cite{martino:survey}. 
In this condition, automatic fake news detection has been developing rapidly.
According to \citeauthor{Ruffo:survey} \shortcite{Ruffo:survey}, automatic fake news detection mainly include textual-content based methods \cite{ Giachanou:textual-content, Bilal:textual-content,Kaliyar:textual-content}, user-role based methods \cite{Vo:user-role, Giachanou:user-role}, multi-modal approaches \cite{Zlatkova:multimodal,Yi:multimodal} and detection of bots and trolls \cite{Stella:bots,Mohsen:bots}.
Among textual-content based methods, fact extraction and verification (FEVER) \cite{thorne:fever}  has been attracting developing attention. 
As shown in  Figure \ref{examples}, for a given claim, the task aims to select at most 5 most related sentences as evidences from millions of open-domain Wikipedia documents for fact extraction, and combine the selected evidences to judge the claim as SUPPORTS, REFUTES or NOT ENOUGH INFO (NEI) for fact verification.

Recently, several strong models \cite{Nie:emnlp,Nie:aaai,zhou:gear,liu:kgat,hidey:dese,subramanian:hesm} have been proposed for the task.
Although they have made great progress and obtained excellent performance on the task, we argue that they fail to utilize multi-view contextual information and thus cannot obtain better performance.
Specifically, we define two kinds of context for each evidence sentence, i.e. \textbf{intra-document context} and \textbf{inter-document context}.
Intra-document context consists of the document title and all the other sentences from the same document.
Inter-document context consists of all other evidences which may come from different documents.
Multi-view contextual information is of great importance for fact extraction and verification.
For instance, as shown in Figure \ref{examples}, intra-document context information can help to clarify the relationship between different entities, e.g. ``\textit{Edar Wright}'' and its alias name  ``\textit{Edar Howard Wright}'' in the first evidence, and ``\textit{Gemini}'' and it coreference ``\textit{this sign}'' in the second sentence.
Besides, the two evidence sentences can be regarded as inter-document context of each other, and the information interaction and fusion between them is essential to verify the claim in this multi-hop sample.

To this end, we propose to integrate multi-view contextual information (IMCI) for fact extraction and fact verification, where we introduce the multi-view contextual information to encode the evidence sentences to handle the task.
In summary, our contributions are as follows:

$\bullet$ We propose an iterative multi-view fact extraction model. It retrieves related documents and extracts related evidence sentences in two iterations, with multi-view context information joined.

$\bullet$ We propose a multi-view fact verification model. Each evidence sentence is encoded from two views, and a dual evidence fusion graph is adopted to fuse the information from diverse views and different evidences.

$\bullet$ Our IMCI framework makes great progress on both fact extraction and verification, and achieves state-of-the-art performance with a winning FEVER score of $72.97\%$ on the online blind test set.

\section{Iterative Multi-view Fact Extraction}

Our fact extraction model iteratively conducts document retrieval and sentence retrieval in two iterations to obtain corresponding candidate evidence sentences, and then reranks the candidates of different iterations for better performance.

\subsection{Document Retrieval}

Document retrieval includes \textit{coarse document retrieval} in iteration 1 and \textit{refined document retrieval} in iteration 2.

Coarse document retrieval aims to quickly obtain most related documents from millions of open-domain Wikipedia documents with as high as possible recall and acceptable precision.
Inspired by UKP-Athene \cite{Andreas:ukp} and SR-MRS \cite{Nie:emnlp}, coarse document retrieval is a combination of constituency-based Wikipedia search and TF-IDF retrieval.
These two respectively utilize search engine power and statistical word frequency information.
For constituency-based Wikipedia search, we also conduct mention filtering like UKP-Athene \cite{Andreas:ukp}. 
That is, if the title of a document is not explicitly mentioned in the claim, then we consider it as weakly related and remove it. 

Refined document retrieval aims to retrieve documents with improved performance than coarse retrieval, namely higher recall and also higher precision and F1 score.
It adopts dense semantic retrieval and utilizes Wikipedia hyperlinks.
Specifically, in iteration 2, we decide refined candidate documents according to corresponding candidate evidences from iteration 1.
That is, for each claim, all documents which contain at least one candidate evidence will be taken into account.
Furthermore, as top one candidate evidences show pretty high precision (86.11\%, in Table \ref{sentence_retrieval}), we regard them as gold evidence, and take all documents which have hyperlinks with them as refined candidate documents to process multi-hop problem.

\subsection{Sentence Retrieval}
\label{1_1}

Sentence retrieval aims at selecting most related sentences as evidences from candidate documents. 
In previous models, during sentence retrieval, it is required to design sampling strategy to obtain negative samples for neural retrieval model training.
Besides, these models respectively encode and score each claim-sentence pair. 

\begin{figure}[h]
\centerline{\includegraphics[scale=0.095]{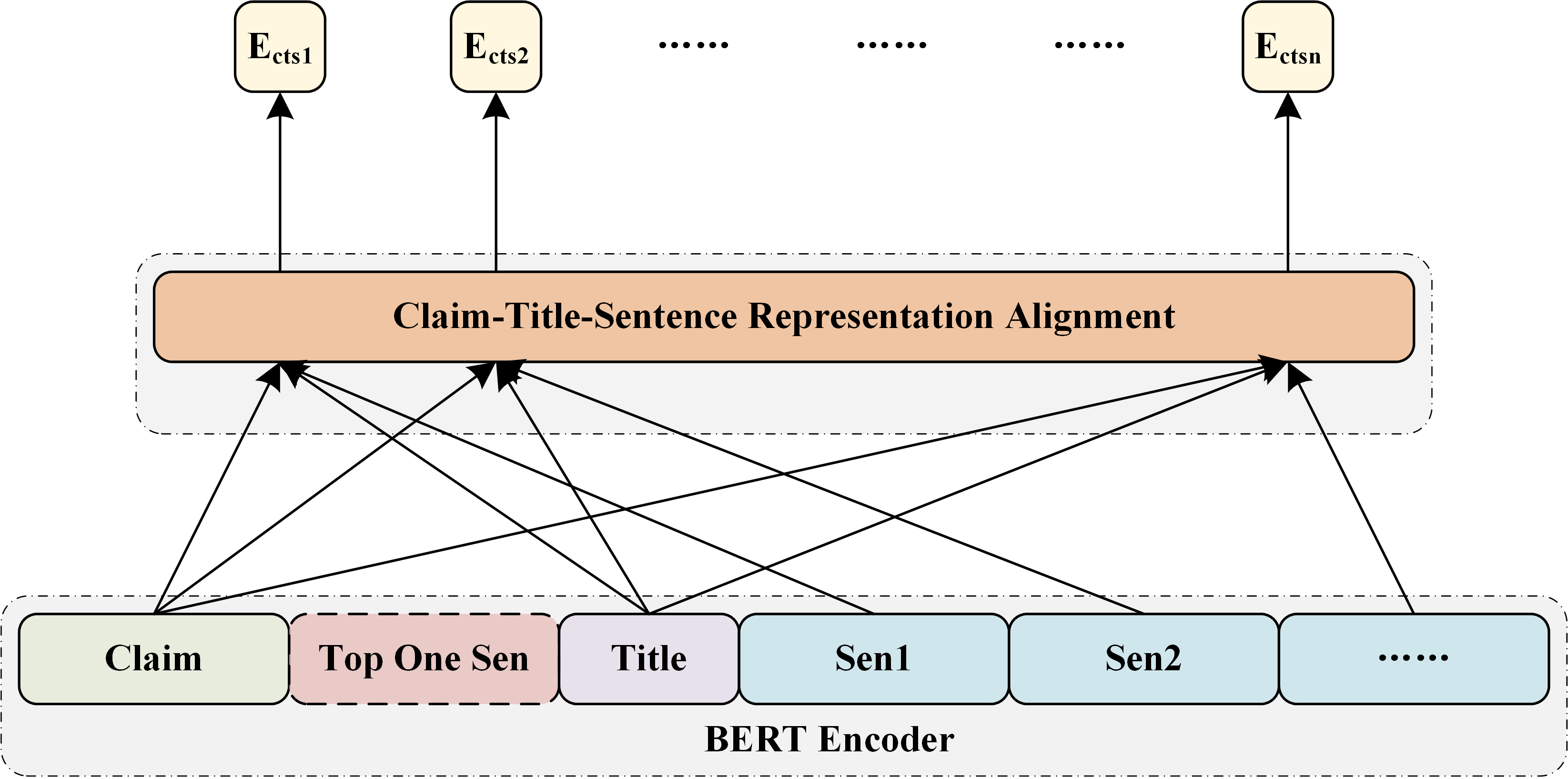}}
\caption{Sentence retrieval model. Sentences are encoded within intra-document context. In iteration 2, we insert top one candidate evidence (red dashed box) into the input sequence as inter-document context.}
\label{sentence retrieval model}
\end{figure}

Differently, in our framework, to avoid sampling strategy design and also utilize multi-view contextual information, we encode each sentence within its corresponding intra-document context.
Moreover, as mentioned, top one candidate evidences of iteration 1 show pretty high precision (86.11\%, in Table \ref{sentence_retrieval}).
Therefore, in iteration 2 we take them as inter-document context, and insert them into the input sequence to process multi-hop problem. 

Formally, as shown in Figure \ref{sentence retrieval model}, the claim, the document title and all the sentences in the document are concatenated: 
\begin{equation}
\begin{split}
[CLS]\;\textit{claim}\;[SEP]\;\textit{sen$^*$}\;[SEP]\;\textit{title}\\
[SEP]\;\textit{sen1}\;\textit{sen2}\;\dots\;[SEP]
\end{split}
\end{equation}
where \textit{sen$^*$} denotes top one candidate evidence of iteration 1, which are taken as inter-document context in iteration 2. 
The sequence is encoded by BERT encoder.
For the claim, we take the hidden state of the first claim token as claim representation \textit{$E_c$}.
For the title, we take the hidden state of the first title token as title representation \textit{$E_t$}.
For each sentence, we take the hidden state of the first sentence token as the sentence representation \textit{$E_s$}.
The sentence representation is enhanced through alignment with the title representation:
\begin{equation}
E_{ts} = W_{a}[E_t, E_s, E_t - E_s, E_t \odot E_s]
\end{equation}
 and the claim representation:
\begin{equation}
E_{cts} = W_{a}^{'}[E_c, E_{ts}, E_c - E_{ts}, E_c \odot E_{ts}]
\end{equation}
where $\odot$ means element-wise Hadamard product.
Then, the score of sentence $\hat{y}$ is obtained through a Multi Layer Perceptron (MLP) with sigmoid activation function:
\begin{equation}
\hat{y} = {\rm Sigmoid} ({\rm MLP}(E_{cts}))
\end{equation}

The training objective of sentence retrieval is defined as binary cross entropy loss, to maximize the probability of groundtruth evidence sentences:
\begin{equation}
\begin{split}
\mathcal{L}_{E} = - \frac {1} {\sum \limits_{i=1}^m n_i}  \sum \limits_{i=1}^{m} \sum \limits_{j=1}^{n_i} [y_{ij} \cdot \log (\hat{y}_{ij}) \\
 + (1 - y_{ij}) \cdot \log (1 - \hat{y}_{ij})]
\end{split}
\label{5}
\end {equation}
where \textit{m} is the batch size, \textit{$n_i$} is the sentence number of document \textit{i}, and \textit{y} is the sentence label, 1 for groundtruth evidence sentences while 0 for non-evidence sentences.

\subsection{Full Pipeline}
In each iteration, we have scored different sentences as candidate evidences.
To obtain better performance, for each claim, we merge the results of different iterations, and rerank the sentences through their scores.
Finally, according to the original setup of the task, we keep at most top 5 sentences as evidences, for further fact verification.

\section{Multi-view Fact Verification}

\subsection{Multi-view Contextual Encoding}

For each evidence sentence, we respectively obtain its representations through intra-document encoding and inter-document encoding.

$\bullet$ \textbf{Intra-document Encoding} aims to capture intra-document contextual information of each evidence sentence.
It is similar to the sentence retrieval model in Section \ref{1_1}.
Each evidence sentence is encoded within its intra-document context.
Then its intra-document representation is also obtained through alignment.

$\bullet$ \textbf{Inter-document Encoding} is utilized to capture token-level information interaction among different evidence sentences to handle multi-hop problem.
The claim, all evidence sentences and their document titles are concatenated as another input sequence:
\begin{equation}
\begin{split}
[CLS]\;\textit{claim}\;[SEP]\;\textit{title1}\;[SEP]\;\textit{evi1}\;[SEP]\\
\textit{title2}\;[SEP]\;\textit{evi2}\;[SEP]\;\textit{$\cdots$}\;[SEP]
\end{split}
\end{equation}
The concatenation is also encoded by BERT encoder.
Then, similarly, we obtain claim, title or evidence representation from the hidden state of the first token.
Finally, for each evidence, we obtain its inter-document representation through alignment with the claim representation and its corresponding title representation.

\subsection{Dual Evidence Fusion Graph}

\begin{figure}[!h]
\centerline{\includegraphics[scale=0.12]{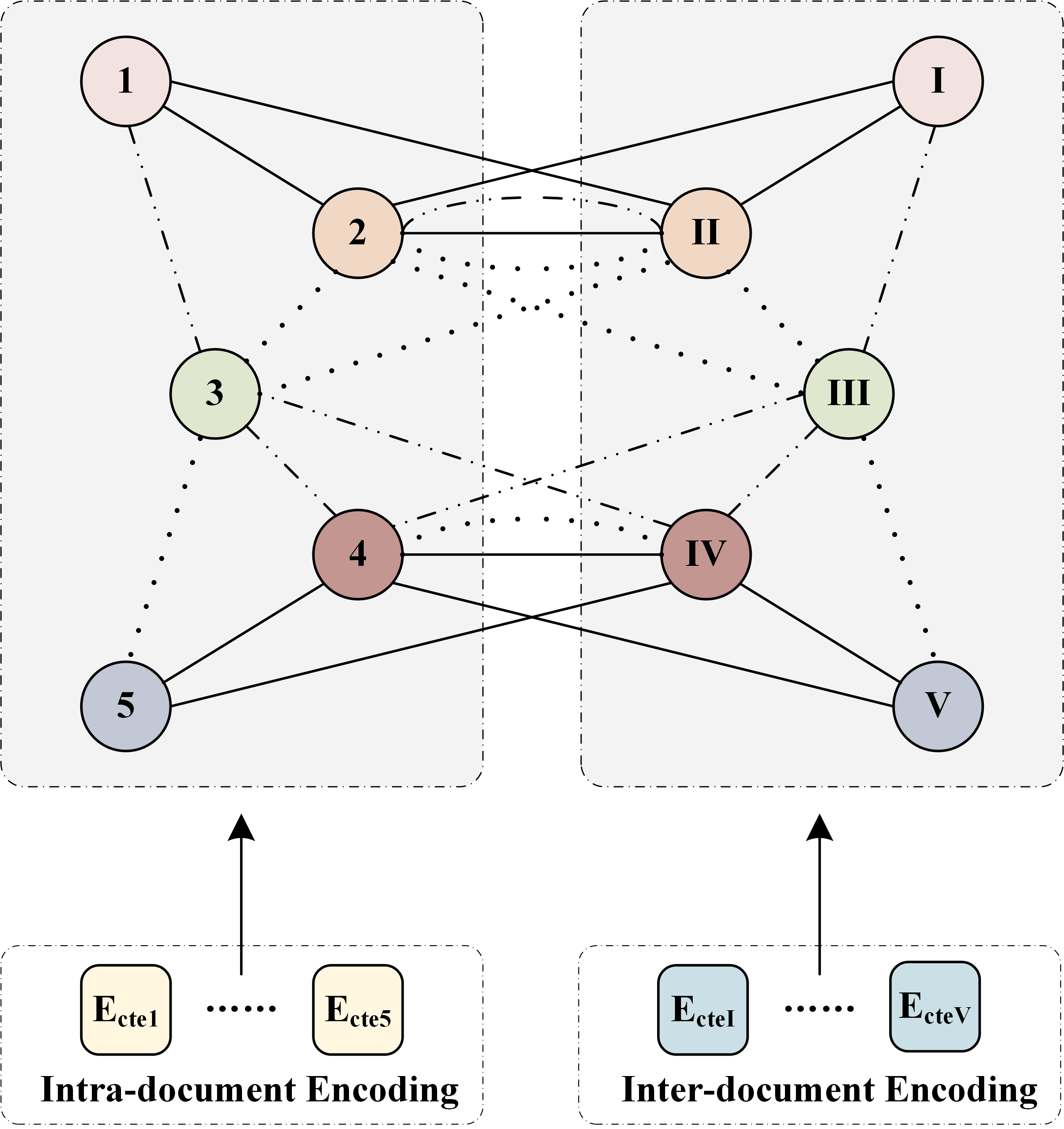}}
\caption{Dual Evidence Fusion Graph Network. Node 1 and node I denote different representations of the same evidence sentence from different encoding methods, similarly for node 2 and node II, etc. We define three kinds of edges in total.}
\label{dual_graph}
\end{figure}

Through multi-view contextual encoding, for each evidence, we can obtain two alignment representations from different contextual views.
To further integrate multi-view evidence information to handle multi-hop problem, inspired by multi-relational graph convolutional network \cite{cao:graph1,tu:graph1,tu:graph2}, we propose dual evidence fusion graph network.
As shown in Figure \ref{dual_graph}, \textit{one evidence sentence} corresponds to \textit{two different nodes} in this graph, whose initial representations respectively come from intra-document encoding and inter-document encoding.
For each evidence sentence,  the noun phrases and named entities are extracted as keywords through spaCy\footnote{\url{https://spacy.io/}} tool.
Then for a pair of nodes, the links between them are decided according to following rules:

$\bullet$ \textbf{Common Document}\;Two nodes are linked if they come from the same document.

$\bullet$ \textbf{Common Keyword}\;Two nodes are linked if they share overlapped keywords.

$\bullet$ \textbf{Claim Jump}\;Two nodes are linked if they respectively share overlapped keywords with the claim. 

For each claim, \textit{N} selected evidence sentences introduce 2\textit{N} evidence nodes.
Let $H_i$ $\in$ $\mathcal{R}^{2N\times d}$ denotes the node representations at \textit{i}-th graph layer, where $d$ refers to the hidden dimension.
The initial representation $H_0$ is the claim-title-evidence alignment representation through multi-view contextual encoding.
Updated information $U_i$ $\in$ $\mathcal{R}^{2N\times d}$ after a single graph layer is defined as :
\begin{equation}
U_i = H_iW_{0} + \sum \limits_{j=1}^{3} \tilde{A}_jH_iW_j
\end{equation}
where $\tilde{A}_j$ $\in$ $\mathcal{R}^{2N\times 2N}$ denotes corresponding row normalized adjacent matrix for different kinds of edges.
Then the forget ratio $G_i$ $\in$ $\mathcal{R}^{2N\times d} $between the updated and old information is:
\begin{equation}
G_i = {\rm Sigmoid}(W_{g}[U_i, H_i])
\label{12}
\end{equation}
And the updated evidence representation through the graph layer is:
\begin{equation}
H_{i+1} = {\rm Activation}(U_i) \odot G_i + H_{i} \odot (1 - G_i)
\label{13}
\end{equation}
In this way, with several stacked layers, evidence representations are updated and multi-view evidence information is fused.

\subsection{Confidence Aggregation}

Aggregation aims to combines the evidence representations for final inference representation to verify the claim.
Among the selected evidence sentences of a claim, some are groundtruth ones while others are not.
To utilize evidence label information to enhance fact verification, like \citeauthor{tu:graph2} \shortcite{tu:graph2}, we adopt confidence aggregation.

Formally, let $H_k$ $\in$ $\mathcal{R}^{2N\times d}$ denotes evidence representations at the last graph layer.
The confidence score of \textit{j}-th evidence $\hat{y}_j$ is obtained from its representation $H_k^j$:
\begin{equation}
\hat{y}_j = {\rm Sigmoid}({\rm MLP}(H_{k}^{j}))
\end{equation}
The final inference representation for fact verification $R_{v}$ is the weighted sum of the evidence representations, where the weights are corresponding confidence scores:
\begin{equation}
R_{v} = \sum \limits_{j=1}^{N} \hat{y}_jH_{k}^{j}
\end{equation}
and the fact verification result is obtained through a 3-way classification network:
\begin{equation}
\hat{v} = {\rm Softmax}(WR_{v}+b)
\end{equation}

The total loss consists of the binary cross entropy loss of evidence confidence, and the cross entropy loss of 3-way fact verification:
\begin{equation}
\mathcal{L}_{\uppercase\expandafter{\romannumeral2}} = BCE(y,\hat{y}) + CE(v,\hat{v})
\end{equation}
Here \textit{y} is the evidence sentence label, 1 for groundtruth evidence sentences and 0 for non-evidence sentences. Besides, \textit{v} is the fact verification label.

\section{Experiment}

\subsection{Dataset}
We conduct our experiments on FEVER 1.0 shared task \cite{thorne:fever}, which consists of 185,455 annotated claims with 5,416,537 Wikipedia documents from the June 2017 dumps. 
We adopt the original dataset split of the task, which includes a training set, a development set and an online blind test set. % \footnote{The leaderboard is on \url{https://competitions.codalab.org/competitions/18814}}.
The detailed information is shown in Table \ref{dataset}.

\setlength\tabcolsep{3pt}
\begin{table}[!h]
\centering
\begin{tabular}{ccccc}
\hline
\textbf{Split} & \textbf{SUPPORTS} & \textbf{REFUTES} & \textbf{NEI} & \textbf{Total}   \\ \hline
train & 80035    & 29775   & 35639           & 145449 \\
dev   & 6666     & 6666    & 6666            & 19998   \\
test  & 6666     & 6666    & 6666            & 19998   \\ \hline
\end{tabular}
\caption{Statistics information of FEVER 1.0 Shared Task.} % (\textit{dev} is short for \textit{development})}
\label{dataset}
\end{table}

Moreover, for a claim, there exist several groups of evidences, and each group itself is enough to independently verify the claim.
To further study the impact of multi-view contextual information, we conduct a refined split on the development set.
Specifically, samples of the development set can be divided into 5 parts and the ratio of different parts are displayed in Table \ref{ratio}:

$\bullet$ \textbf{Single}. All evidence groups contain exactly one sentence.

$\bullet$ \textbf{Single+}. At least one evidence group contains only one sentence, and at least one group contains multi sentences.

$\bullet$ \textbf{Multi}. All evidence groups contain exactly two sentences.

$\bullet$ \textbf{Multi+}. All evidence groups contain multi sentences, and at least one group contains more than two sentences.

$\bullet$ \textbf{NEI}. The sample is labeled as NEI with no evidence groups annotated.

%The ratio of different parts are displayed in Table \ref{ratio}.
%Single-hop ones (Single, Single+) take about 60.65\%, multi-hop ones (Multi, Multi+) take about 6.02\%, while NEI ones take about 33.33\%.

\begin{table}[!h]
\centering
\begin{tabular}{ccccc}
\hline
\textbf{Single} & \textbf{Single+} & \textbf{Multi} & \textbf{Multi+} & \textbf{NEI} \\ \hline
56.87           & 3.78             & 5.03           & 0.99            & 33.33        \\ \hline
\end{tabular}
\caption{Ratio of different parts on the development set.}
\label{ratio}
\end{table}

\subsection{Experiment Setup}

Our IMCI is implemented through Pytorch 1.2.0 and our experiments are conducted on a computation node with 4 NVIDIA Titan V GPU.
Pre-trained BERT \cite{devlin:bert} encoder is employed for all experiments.
We also try RoBERTa encoder~\cite{yinhan:roberta}  for fact verification.
For the claims, we set max length as 64, and claims longer than this will be truncated.
For the encoders, we set max input sequence length as 512, and sequence longer than this will be split with stride window size of 128.
We utilize BERTAdam optimizer with initial learning rate of 1e-5 and warmup ratio of 0.1.
For sentence retrieval, we adopt mini batch size of 4 and gradient accumulation step of 8.
In each iteration, we train 2 epochs and select top 5 sentences as candidate evidences.
For fact verification, we adopt mini batch size of 1 and gradient accumulation step of 32.
For dual evidence graph, we stack 3 graph layers, where the hidden dimension is the same as that of the encoder.
In each condition, we randomly start 4 times, train 4 epochs, and choose model parameters with the best performance on the development set.

\subsection{Evaluation Metric}
We adopt FEVER score as the dominant evaluation metric, which is the officially chief metric.
FEVER score requests that fact verification label is correctly predicted, and at least one complete group of evidence sentences is found for SUPPORTS and REFUTES samples.
The second important metric is label accuracy.
For document retrieval and sentence retrieval, we take precision, recall as well as F1 into account.
Here, we attach more importance to recall according to the task setting.

\section{Results}

\subsection{Main Results}

Main results on the blind test set are shown in Table \ref{final test}.
With multi-view contextual information joined, our ICMI framework obtains FEVER score of 70.10\%  and label accuracy of 73.04\% with BERT$_{base}$ encoder.
The performance is comparable and even slightly promoted compared with the state-of-the-art one among all baselines with BERT$_{base}$ encoder.
Moreover, our model with RoBERTa$_{base}$ encoder obtains FEVER score of 72.97\% and label accuracy of 75.84\%, and shows even higher performance than several baselines with large encoder.
% Furthermore, our model with RoBERTa$_{large}$ encoder obtains FEVER score of 74.42\% and label accuracy of 77.38\%, and significantly outperforms all baselines.
These indicate that our framework has made great progress to conduct more accurate fact extraction and verification.

\setlength\tabcolsep{1pt}
\begin{table}[!h]
\centering
\begin{tabular}{cccc}
\hline
\textbf{Model}          & \textbf{LA} & \textbf{FEVER} \\ \hline
UKP-Athene\citeyearpar{Andreas:ukp}                      & 65.46          & 61.58 \\
QFE\citeyearpar{Nishida}                     &69.30   &61.80 \\
NSMN\citeyearpar{Nie:aaai}                        & 68.16          & 64.23 \\ \hline
GEAR-BERT$_{base}$\citeyearpar{zhou:gear}         & 71.60          & 67.10 \\
SR-MRS-BERT$_{base}$\citeyearpar{Nie:emnlp}       & 72.56          & 67.26 \\
DeSePtion-BERT$_{base}$\citeyearpar{hidey:dese}    & 72.47          & 68.80 \\
Transformer-XH-BERT$_{base}$\citeyearpar{zhao:xh} & 72.39        & 69.07 \\
KGAT-BERT$_{base}$\citeyearpar{liu:kgat}         & 72.81          & 69.40 \\
CorefBERT-BERT$_{base}$\citeyearpar{ye:corefbert}    & 72.88          & 69.82 \\
HESM-ALBERT$_{base}$\citeyearpar{subramanian:hesm}       & \textbf{73.25}          & 70.06 \\
HESM-BERT$_{base}$\citeyearpar{subramanian:hesm}         & 73.18          & 70.07 \\ \hline
\textit{ours} IMCI-BERT$_{base}$     & 73.04         & \textbf{70.10} \\ \hline \hline
KGAT-BERT$_{large}$\citeyearpar{liu:kgat}        & 73.61          & 70.24 \\
CorefBERT-BERT$_{large}$\citeyearpar{ye:corefbert}   & 74.37          & 70.86 \\
HESM-ALBERT$_{large}$\citeyearpar{subramanian:hesm}      & 74.64          & 71.48 \\
KGAT-RoBERTa$_{large}$\citeyearpar{liu:kgat}     & 74.07          & 70.38 \\
CorefBERT-RoBERTa$_{large}$\citeyearpar{ye:corefbert}      &\textbf{75.96}          & 72.30 \\ \hline
% MLA-RoBERTa$_{large}$\citeyearpar{xin:mla} &77.05 &73.72 \\ \hline
\textit{ours} IMCI-RoBERTa$_{base}$      &75.84     &\textbf{72.97}  \\ \hline
% \textit{ours} IMCI-RoBERTa$_{large}$      &\textbf{77.38}     &\textbf{74.42}  \\ \hline

\end{tabular}
\caption{Overall performance on the online blind test det. FEVER is the officially chief score. LA denotes label accuracy.}
\label{final test}
\end{table}

\subsection{Document Retrieval}

Document retrieval results of different iterations on the development set are displayed in Table \ref{document_retrieval}.
With search engine power adopted through Wikipedia search, and statistical word frequency information joined through TF-IDF retrieval, coarse document retrieval obtains the highest recall of 92.77\%.
Besides, with dense semantic retrieval model guided and top one evidence hyperlinks joined, refined document retrieval shows even higher recall of 95.69\%.
Furthermore, refined document retrieval has much higher precision of 29.90\% and F1 of 45.56\%, respectively obtains 21.22\% and 29.69\% absolute increase than coarse retrieval.
Therefore, our iterative fact extraction model has made great improvement on document retrieval.

\setlength\tabcolsep{3pt}
\begin{table}[]
\centering
\begin{tabular}{cccc}
\hline
\textbf{Model}    & \textbf{P} & \textbf{R}                      & \textbf{F1} \\ \hline
UKP-Athene\citeyearpar{Andreas:ukp}        & -          & 90.32                       & -           \\
NSMN\citeyearpar{Nie:aaai}              & \textbf{51.04}      & 89.23   & \textbf{64.94}       \\
GEAR-BERT$_{base}$\citeyearpar{zhou:gear}              & -          & 89.99                       & -           \\
SR-MRS-BERT$_{base}$\citeyearpar{Nie:emnlp}           & 18.11      & 92.03                           & 30.27       \\ \hline
Coarse Retrieval  & 8.68       & \textbf{92.77}                 & 15.87       \\
Refined Retrieval & 29.90      & \textbf{95.69}                 & 45.56       \\ \hline
\end{tabular}
\caption{Document retrieval results on the development set. - denotes that the item is not available.}
\label{document_retrieval}
\end{table}

\begin{figure}[h]
\centerline{\includegraphics[scale=0.48]{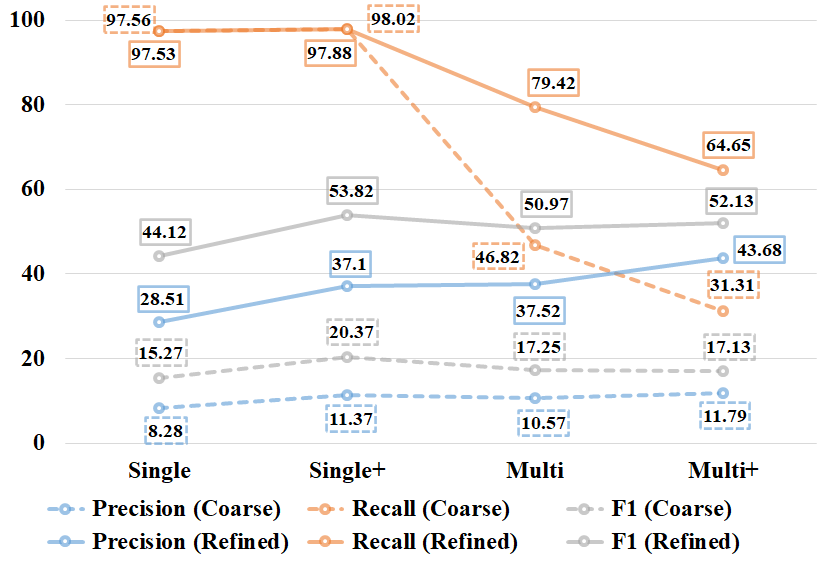}}
\caption{Document retrieval results on different parts of the development set. NEI samples are not taken into consideration since no evidence sentences are annotated for them.}
\label{document_retrieval_parts}
\end{figure}

Moreover, document retrieval results on different parts of the development set are displayed in Figure \ref{document_retrieval_parts}. 
Coarse document retrieval can handle Single and Single+ samples, where the recall are respectively as high as 97.56\% and 98.02\%.
However, coarse document retrieval fails to handle multi-hop samples, where the recall of Multi and Multi+ samples are both pretty low, respectively 46.82\% and 31.31\%.
Compared to coarse document retrieval, refined document retrieval shows comparable recalls but much higher precision and F1 score on Single and Single+ samples.
Furthermore, refined document retrieval makes great progress on multi-hop samples.
For Multi and Multi+ samples, refined document retrieval respectively achieves 32.60\% and 33.34\% absolute increase on recall.
Besides, the precision and F1 score also get significantly improved.
These results indicate the high efficiency of our iterative multi-view fact extraction model.
However, although refined document retrieval has achieved significant improvement, the recall of multi-hop samples is still far lower than single-hop ones.

\subsection{Sentence Retrieval}

Sentence retrieval results on the development set are summarized in Table \ref{sentence_retrieval}.
Our IMCI framework obtains the highest recall of 92.86\%, and significantly outperforms all baselines.

\setlength\tabcolsep{3pt}
\begin{table}[!h]
\centering
\begin{tabular}{cccc}
\hline
\textbf{Model}     & \textbf{P} & \textbf{R} & F1    \\ \hline
UKP-Athene\citeyearpar{Andreas:ukp}         & -                  & 86.24           & -     \\
NSMN\citeyearpar{Nie:aaai}              & 36.49              & 86.79           & 51.38 \\
GEAR-BERT$_{base}$\citeyearpar{zhou:gear}               & 24.08              & 86.72           & 37.69 \\
SR-MRS-BERT$_{base}$\citeyearpar{Nie:emnlp}             & 44.47              & 86.60           & 58.77 \\
HESM-BERT$_{base}$\citeyearpar{subramanian:hesm}$^\#$              & -                  & 90.50           & -     \\ \hline
Iteration 1        & 25.31              & \textbf{90.30}           & 39.54 \\
w.o. Alignment     & 24.86              & 90.16           & 38.97 \\
Iteration 1 (Top 1) & \textbf{86.11}              & 78.08           & \textbf{81.90} \\ \hline
Iteration 2        & 25.90              & \textbf{91.98}           & 40.42 \\ \hline
IMCI$^\#$               & 25.74              & \textbf{92.86}           & 40.30 \\ \hline
\end{tabular}
\caption{Sentence retrieval results on the development set. According to the original task setup, we keep top 5 sentences as evidence for each claim. $\#$ means the models adopt iterative sentence retrieval. \textit{w.o.} means without the item.}
\label{sentence_retrieval}
\end{table}

For sentence retrieval of iteration 1, upstream coarse document retrieval obtains an extremely low precision of 8.68\% (in Table \ref{document_retrieval}).
Thus, for each claim, on average our sentence retrieval model is requested to distinguish top 5 sentences as candidate evidences from more than 250 sentences.
In this condition, with intra-document contextual information joined, the model obtains pretty high recall of 90.30\%, and shows comparable performance with state-of-the-art iterative sentence retrieval model HESM\cite{subramanian:hesm}.
Besides, top one candidate evidences show a pretty high precision of 86.11\%.
This shows the importance of intra-document context, and is the base of refined document retrieval.
Besides, the high precision also guarantees that top one candidate evidences can be considered as inter-document context in sentence retrieval of iteration 2.
With multi-view contextual information joined, our sentence retrieval model of iteration 2 obtains even higher recall of 91.98\%.
Moreover, full pipeline reranking makes the recall get far more increase to 92.86\%.
These show the great power of multi-view contextual information on fact extraction.

Moreover, sentence retrieval results on different parts of the development set are displayed in detail in Table \ref{detailed_sentence_retrieval}.
For Single and Single+ samples, the recall are respectively high at 96.33\% and 95.90\%, while the precision and F1 score are pretty low.
However, the recall of Multi samples is pretty low at 64.41\%, while that of Multi+ samples is far lower at 26.26\%.
Therefore, taking these and the document retrieval results in Figure \ref{document_retrieval_parts} into consideration, it seems that fact extraction for multi-hop samples is still a difficult problem, although our model has made several progress. 
% Future fact extraction models should pay more attention to promoting precision for single-hop samples, and recall for multi-hop samples.

\setlength\tabcolsep{6pt}
\begin{table}[htp]
\centering
\begin{tabular}{cccc}
\hline
\textbf{Part} & \textbf{P} & \textbf{R} & \textbf{F1} \\ \hline
Single        & 23.00              & 96.33           & 37.14       \\
Single+       & 52.06              & 95.90           & 67.48       \\
Multi         & 32.98              & 64.41           & 43.63       \\
Multi+        & 45.25              & 26.26           & 33.23       \\ \hline
\end{tabular}
\caption{Sentence retrieval results on different parts of the development set. NEI samples are not taken into consideration since no evidence sentences are annotated for them.}
\label{detailed_sentence_retrieval}
\end{table}

\subsection{Fact Verification}

\begin{figure}[h]
\centerline{\includegraphics[scale=0.4]{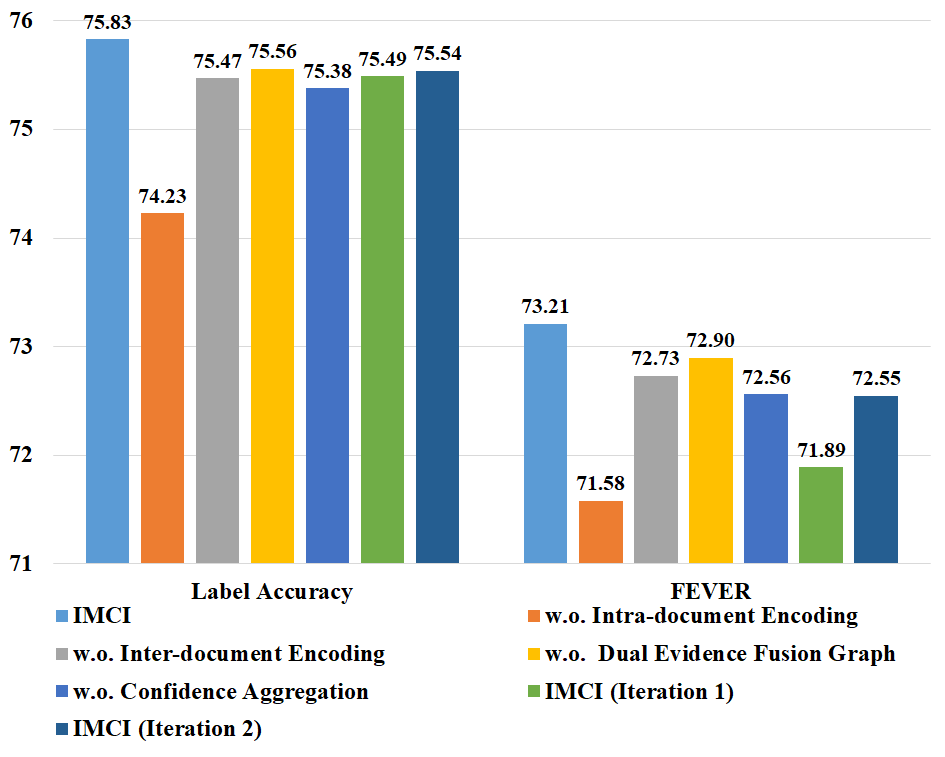}}
\caption{Fact verification results on the development set. These are averaged results on 4 random starts. \textit{w.o.} means without the item.}
\label{fact_verification}
\end{figure}

Fact verification results on the development set are shown in Figure \ref{fact_verification}.
With multi-view contextual information joined, our IMCI framework obtains the highest label accuracy of 75.83\% and the highest FEVER score of 73.21\%.
When ignoring intra-document encoding, the label accuracy and FEVER score suffer severe decrease to 74.23\% and 71.58\%.
This indicates the great importance of intra-document contextual information on fact verification.
However, compared to intra-document encoding, inter-document encoding and dual evidence fusion graph have relatively weak influence on the performance.
These two components mainly aims to handle multi-hop samples.
However, multi-hop samples take pretty low ratio (about 6.02\% in total in Table \ref{ratio}).
Even worse, multi-hop samples have suffered serious performance damage on upstream fact verification task (in Table \ref{sentence_retrieval}).
Evidence confidence aggregation also makes some contribution, indicating the influence of evidence label information.
Besides, we also study the influence of fact verification.
It seems that progress on fact verification mainly contributes to FEVER score while shows weak influence on label accuracy.

\begin{figure}[!h]
\centerline{\includegraphics[scale=0.54]{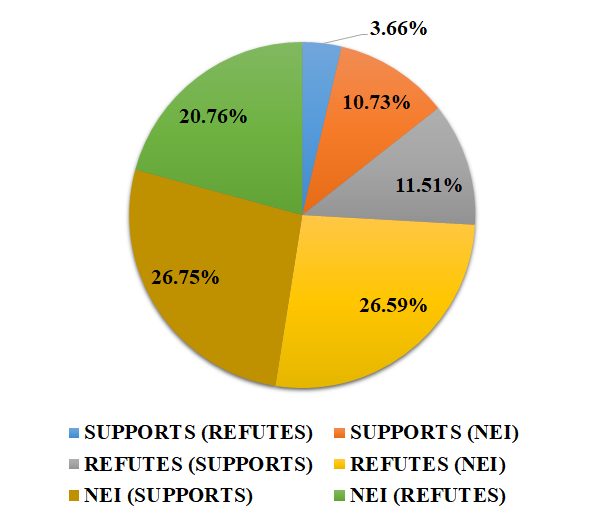}}
\caption{Statistic information of prediction errors on fact verification. (Label \textbf{out of} brackets denotes groundtruth, while label \textbf{in} brackets denotes wrong prediction.)}
\label{error statistic}
\end{figure}

Moreover, for our IMCI framework, the statistic information of prediction errors on fact verification is shown in Figure \ref{error statistic}.
The framework can correctly distinguish SUPPORTS and REFUTES examples, since SUPPORTS (REFUTES) and REFUTES (SUPPORTS) errors respectively take about 3.66\% and 11.51\%.
This may indicate that the logical boundary between SUPPORTS and REFUTES is relatively clear.
Besides, the framework hardly mistakes SUPPORTS examples for NEI examples.
However, it may be difficult for the framework to distinguish REFUTES examples from NEI examples, as well as NEI examples from non-NEI examples, for REFUTES (NEI), NEI (SUPPORTS), and NEI (REFUTES) errors respectively take 26.59\%, 26.75\%, and  20.76\%.
The situation may be due to the pretty unbalanced label distribution of the training set (in Table \ref{dataset}).
Besides, NEI may contain more complex logic semantic than the other two.

\section{Related Work}

$\bullet$ \textbf{Fake News Detection}\;Fake news detection has been attracting more attention. 
\citeauthor{Ruffo:survey} \shortcite{Ruffo:survey} give a detailed survey about the development of this field.
Textual-content based methods \cite{ Giachanou:textual-content, Bilal:textual-content,Kaliyar:textual-content} aim at understanding the linguistic and semantic information in the text to detect fake news.
User-role based methods \cite{Vo:user-role, Giachanou:user-role} pay more attention to the role of users in the propagation of fake news. 
Multi-modal approaches \cite{Zlatkova:multimodal,Yi:multimodal} involve multi-modal information, i.e. text, table, knowledge base, image, speech and video, to evaluate the credibility of news.
Besides, bots and trolls aim at influencing users with commercial, political or ideological purposes by spreading disinformation deliberately.
The detection of them \cite{Stella:bots,Mohsen:bots} is also an important direction.
Moreover, \citeauthor{sheng:zoom} \shortcite{sheng:zoom} recently propose news environment perception for fake news detection, which focus on the background environment of fake news.

\noindent $\bullet$ \textbf{Fact Extraction}\; Fact extraction includes document retrieval and sentence retrieval.
For document retrieval, \citeauthor{Andreas:ukp} \shortcite{Andreas:ukp} propose a constituency-based Wikipedia search model.
\citeauthor{Nie:aaai} \shortcite{Nie:aaai} utilize a keyword matching model based on a quick string matching algorithm \textit{FlashText} \cite{singh}.
\citeauthor{Nie:emnlp} \shortcite{Nie:emnlp} further adopt a combination model of keyword match and TF-IDF retrieval.

For sentence retrieval, \citeauthor{Andreas:ukp} \shortcite{Andreas:ukp}, \citeauthor{Nie:aaai} \shortcite{Nie:aaai}, and \citeauthor{zhou:gear} \shortcite{zhou:gear} respectively modify Enhanced Sequential Inference Model (ESIM) \cite{chen:esim}.
These models separately encode the claim and an evidence sentence, and adopt cross-attention mechanism to accomplish information interaction between the claim and the evidence sentence.
\citeauthor{Nie:emnlp} \shortcite{Nie:emnlp} and  \citeauthor{liu:kgat} \shortcite{liu:kgat} adopt BERT-based model.
\citeauthor{subramanian:hesm} \shortcite{subramanian:hesm} propose iterative fact verification models to retrieve evidence sentences and combine evidence sets.

\noindent $\bullet$ \textbf{Fact Verification}\;For fact verification, \citeauthor{Nie:emnlp} \shortcite{Nie:emnlp} concatenate the claim and the evidence sentences into a sequence as input to BERT encoder, and take the hidden state of the first special token [CLS], as final inference representation.
\citeauthor{zhou:gear} \shortcite{zhou:gear} adopt graph neural network for evidence aggregating and reasoning.
\citeauthor{zhong:dream} \shortcite{zhong:dream} introduce semantic role information to construct refined graph, and adopt graph convolutional network to handle the task.
\citeauthor{liu:kgat} \shortcite{liu:kgat} propose fine-grained kernel-based graph attention network for information interaction between the claim and the evidences.
\citeauthor{subramanian:hesm} \shortcite{subramanian:hesm} propose to combine evidence sets during fact extraction, and conduct fact verification on evidence sets.
\citeauthor{si:topic}  \shortcite{si:topic} introduce topic model and stance detection model, and study the influence of topic and stance information on fact verification.

%All these models do not take multi-view, especially intra-document, contextual information into consideration.
%Differently, our IMCI focus on contextual information to enhance fact extraction and verification.

\section{Conclusion}

In this paper, we propose to integrate multi-view contextual information for fact extraction and verification.
Our experimental results show that our IMCI model can obtain state-of-the-art performance on the task.
Moreover, the ablation study results indicate that multi-view contextual information is essential for both fact extraction and fact verification.
In the future, we will explore much stronger model to utilize contextual information in a more efficient way.

\bibliography{IMCI}

\begin{thebibliography}{33}
\expandafter\ifx\csname natexlab\endcsname\relax\def\natexlab#1{#1}\fi

\bibitem[{Cao et~al.(2019)Cao, Aziz, and Titov}]{cao:graph1}
Nicola~De Cao, Wilker Aziz, and Ivan Titov. 2019.
\newblock \href {https://doi.org/10.18653/v1/n19-1240} {Question answering by
  reasoning across documents with graph convolutional networks}.
\newblock In \emph{Proceedings of the 2019 Conference of the North American
  Chapter of the Association for Computational Linguistics: Human Language
  Technologies}, pages 2306--2317. Association for Computational Linguistics.

\bibitem[{Chen et~al.(2017)Chen, Zhu, Ling, Wei, Jiang, and Inkpen}]{chen:esim}
Qian Chen, Xiaodan Zhu, ZhenHua Ling, Si~Wei, Hui Jiang, and Diana Inkpen.
  2017.
\newblock \href {https://doi.org/10.18653/v1/P17-1152} {Enhanced lstm for
  natural language inference}.
\newblock In \emph{Proceedings of the 55th Annual Meeting of the Association
  for Computational Linguistics}, pages 1657--1668. Association for
  Computational Linguistics.

\bibitem[{Devlin et~al.(2019)Devlin, Chang, Lee, and Toutanova}]{devlin:bert}
Jacob Devlin, Ming~Wei Chang, Kenton Lee, and Kristina Toutanova. 2019.
\newblock \href {https://doi.org/10.18653/v1/n19-1423} {Bert: Pre-training of
  deep bidirectional transformers for language understanding}.
\newblock In \emph{Proceedings of the 2019 Conference of the North American
  Chapter of the Association for Computational Linguistics: Human Language
  Technologies}, pages 4171--4186. Association for Computational Linguistics.

\bibitem[{Fung et~al.(2021)Fung, Thomas, Reddy, Polisetty, Ji, Chang, McKeown,
  Bansal, and Sil}]{Yi:multimodal}
Yi~Fung, Christopher Thomas, Revanth~Gangi Reddy, Sandeep Polisetty, Heng Ji,
  ShihFu Chang, Kathleen~R. McKeown, Mohit Bansal, and Avi Sil. 2021.
\newblock \href {https://doi.org/10.18653/v1/2021.acl-long.133} {Infosurgeon:
  Cross-media fine-grained information consistency checking for fake news
  detection}.
\newblock In \emph{Proceedings of the 59th Annual Meeting of the Association
  for Computational Linguistics and the 11th International Joint Conference on
  Natural Language Processing}, pages 1683--1698. Association for Computational
  Linguistics.

\bibitem[{Ghanem et~al.(2020)Ghanem, Rosso, and Pardo}]{Bilal:textual-content}
Bilal Ghanem, Paolo Rosso, and Francisco M.~Rangel Pardo. 2020.
\newblock \href {https://doi.org/10.1145/3381750} {An emotional analysis of
  false information in social media and news articles}.
\newblock \emph{ACM Transactions on Internet Technology}, 20(2):19:1--19:18.

\bibitem[{Giachanou et~al.(2020)Giachanou, ARissola, Ghanem, Crestani, and
  Rosso}]{Giachanou:user-role}
Anastasia Giachanou, Esteban ARissola, Bilal Ghanem, Fabio Crestani, and Paolo
  Rosso. 2020.
\newblock \href {https://doi.org/10.1007/978-3-030-51310-8\_17} {The role of
  personality and linguistic patterns in discriminating between fake news
  spreaders and fact checkers}.
\newblock In \emph{Proceedings of the 25th International Conference on
  Applications of Natural Language to Information Systems}, Lecture Notes in
  Computer Science, pages 181--192. Springer.

\bibitem[{Giachanou et~al.(2019)Giachanou, Rosso, and
  Crestani}]{Giachanou:textual-content}
Anastasia Giachanou, Paolo Rosso, and Fabio Crestani. 2019.
\newblock \href {https://doi.org/10.1145/3331184.3331285} {Leveraging emotional
  signals for credibility detection}.
\newblock In \emph{Proceedings of the 42nd International ACM SIGIR Conference
  on Research and Development in Information Retrieval}, pages 877--880. ACM.

\bibitem[{Hanselowski et~al.(2018)Hanselowski, Zhang, Li, Sorokin, Schiller,
  Schulz, and Gurevych}]{Andreas:ukp}
Andreas Hanselowski, Hao Zhang, Zile Li, Daniil Sorokin, Benjamin Schiller,
  Claudia Schulz, and Iryna Gurevych. 2018.
\newblock \href {http://arxiv.org/abs/1809.01479} {Ukp-athene: Multi-sentence
  textual entailment for claim verification}.
\newblock \emph{CoRR}, abs/1809.01479.

\bibitem[{Hidey et~al.(2020)Hidey, Chakrabarty, Alhindi, Varia, Krstovski,
  Diab, and Muresan}]{hidey:dese}
Christopher Hidey, Tuhin Chakrabarty, Tariq Alhindi, Siddharth Varia, Kriste
  Krstovski, Mona~T. Diab, and Smaranda Muresan. 2020.
\newblock \href {https://doi.org/10.18653/v1/2020.acl-main.761} {Deseption:
  Dual sequence prediction and adversarial examples for improved
  fact-checking}.
\newblock In \emph{Proceedings of the 58th Annual Meeting of the Association
  for Computational Linguistics}, pages 8593--8606. Association for
  Computational Linguistics.

\bibitem[{Kaliyar et~al.(2021)Kaliyar, Goswami, and
  Narang}]{Kaliyar:textual-content}
Rohit~Kumar Kaliyar, Anurag Goswami, and Pratik Narang. 2021.
\newblock \href {https://doi.org/10.1007/s11042-020-10183-2} {Fakebert: Fake
  news detection in social media with a bert-based deep learning approach}.
\newblock \emph{Multimedia Tools and Applications}, 80(8):11765--11788.

\bibitem[{Kruengkrai et~al.(2021)Kruengkrai, Yamagishi, and Wang}]{xin:mla}
Canasai Kruengkrai, Junichi Yamagishi, and Xin Wang. 2021.
\newblock \href {https://doi.org/10.18653/v1/2021.findings-acl.217} {A
  multi-level attention model for evidence-based fact checking}.
\newblock In \emph{Findings of the 59th Association for Computational
  Linguistics1}, pages 2447--2460. Association for Computational Linguistics.

\bibitem[{Liu et~al.(2019)Liu, Ott, Goyal, Du, Joshi, Chen, Levy, Lewis,
  Zettlemoyer, and Stoyanov}]{yinhan:roberta}
Yinhan Liu, Myle Ott, Naman Goyal, Jingfei Du, Mandar Joshi, Danqi Chen, Omer
  Levy, Mike Lewis, Luke Zettlemoyer, and Veselin Stoyanov. 2019.
\newblock \href {http://arxiv.org/abs/1907.11692} {Roberta: A robustly
  optimized {BERT} pretraining approach}.
\newblock \emph{CoRR}, abs/1907.11692.

\bibitem[{Liu et~al.(2020)Liu, Xiong, Sun, and Liu}]{liu:kgat}
Zhenghao Liu, Chenyan Xiong, Maosong Sun, and Zhiyuan Liu. 2020.
\newblock \href {https://doi.org/10.18653/v1/2020.acl-main.655} {Fine-grained
  fact verification with kernel graph attention network}.
\newblock In \emph{Proceedings of the 58th Annual Meeting of the Association
  for Computational Linguistics}, pages 7342--7351. Association for
  Computational Linguistics.

\bibitem[{Martino et~al.(2020)Martino, Cresci, Cedeno, Yu, Pietro, and
  Nakov}]{martino:survey}
Giovanni Da~San Martino, Stefano Cresci, Alberto~Barron Cedeno, Seunghak Yu,
  Roberto~Di Pietro, and Preslav Nakov. 2020.
\newblock \href {https://doi.org/10.24963/ijcai.2020/672} {A survey on
  computational propaganda detection}.
\newblock In \emph{Proceedings of the 29th International Joint Conference on
  Artificial Intelligence}, pages 4826--4832. ijcai.org.

\bibitem[{Nie et~al.(2019{\natexlab{a}})Nie, Chen, and Bansal}]{Nie:aaai}
Yixin Nie, Haonan Chen, and Mohit Bansal. 2019{\natexlab{a}}.
\newblock \href {https://doi.org/10.1609/aaai.v33i01.33016859} {Combining fact
  extraction and verification with neural semantic matching networks}.
\newblock In \emph{Proceedings of the 33rd AAAI Conference on Artificial
  Intelligence}, pages 6859--6866. AAAI Press.

\bibitem[{Nie et~al.(2019{\natexlab{b}})Nie, Wang, and Bansal}]{Nie:emnlp}
Yixin Nie, Songhe Wang, and Mohit Bansal. 2019{\natexlab{b}}.
\newblock \href {https://doi.org/10.18653/v1/D19-1258} {Revealing the
  importance of semantic retrieval for machine reading at scale}.
\newblock In \emph{Proceedings of the 2019 Conference on Empirical Methods in
  Natural Language Processing and the 9th International Joint Conference on
  Natural Language Processing}, pages 2553--2566. Association for Computational
  Linguistics.

\bibitem[{Nishida et~al.(2019)Nishida, Nishida, Nagata, Otsuka, Saito, Asano,
  and Tomita}]{Nishida}
Kosuke Nishida, Kyosuke Nishida, Masaaki Nagata, Atsushi Otsuka, Itsumi Saito,
  Hisako Asano, and Junji Tomita. 2019.
\newblock \href {https://doi.org/10.18653/v1/p19-1225} {Answering while
  summarizing: Multi-task learning for multi-hop qa with evidence extraction}.
\newblock In \emph{Proceedings of the 57th Conference of the Association for
  Computational Linguistics}, pages 2335--2345. Association for Computational
  Linguistics.

\bibitem[{Ruffo et~al.(2021)Ruffo, Semeraro, Giachanou, and
  Rosso}]{Ruffo:survey}
Giancarlo Ruffo, Alfonso Semeraro, Anastasia Giachanou, and Paolo Rosso. 2021.
\newblock \href {https://arxiv.org/abs/2109.07909} {Surveying the research on
  fake news in social media: a tale of networks and language}.
\newblock \emph{CoRR}, abs/2109.07909.

\bibitem[{Sayyadiharikandeh et~al.(2020)Sayyadiharikandeh, Varol, Yang,
  Flammini, and Menczer}]{Mohsen:bots}
Mohsen Sayyadiharikandeh, Onur Varol, KaiCheng Yang, Alessandro Flammini, and
  Filippo Menczer. 2020.
\newblock \href {https://doi.org/10.1145/3340531.3412698} {Detection of novel
  social bots by ensembles of specialized classifiers}.
\newblock In \emph{Proceedings of 29th ACM International Conference on
  Information and Knowledge Management}, pages 2725--2732. {ACM}.

\bibitem[{Sheng et~al.(2022)Sheng, Cao, Zhang, Li, Wang, and Zhu}]{sheng:zoom}
Qiang Sheng, Juan Cao, Xueyao Zhang, Rundong Li, Danding Wang, and Yongchun
  Zhu. 2022.
\newblock \href {https://doi.org/10.48550/arXiv.2203.10885} {Zoom out and
  observe: News environment perception for fake news detection}.
\newblock \emph{CoRR}, abs/2203.10885.

\bibitem[{Si et~al.(2021)Si, Zhou, Li, Shi, and He}]{si:topic}
Jiasheng Si, Deyu Zhou, Tongzhe Li, Xingyu Shi, and Yulan He. 2021.
\newblock \href {https://doi.org/10.18653/v1/2021.acl-long.128} {Topic-aware
  evidence reasoning and stance-aware aggregation for fact verification}.
\newblock In \emph{Proceedings of the 59th Annual Meeting of the Association
  for Computational Linguistics}, pages 1612--1622. Association for
  Computational Linguistics.

\bibitem[{Singh(2017)}]{singh}
Vikash Singh. 2017.
\newblock \href {http://arxiv.org/abs/1711.00046} {Replace or retrieve keywords
  in documents at scale}.
\newblock \emph{CoRR}, abs/1711.00046.

\bibitem[{Stella et~al.(2018)Stella, Ferrara, and Domenico}]{Stella:bots}
Massimo Stella, Emilio Ferrara, and Manlio~De Domenico. 2018.
\newblock \href {https://doi.org/10.1073/pnas.1803470115} {Bots increase
  exposure to negative and inflammatory content in online social systems}.
\newblock \emph{Proceedings of the National Academy of Sciences},
  115(49):12435--12440.

\bibitem[{Subramanian and Lee(2020)}]{subramanian:hesm}
Shyam Subramanian and Kyumin Lee. 2020.
\newblock \href {https://doi.org/10.18653/v1/2020.emnlp-main.627} {Hierarchical
  evidence set modeling for automated fact extraction and verification}.
\newblock In \emph{Proceedings of the 2020 Conference on Empirical Methods in
  Natural Language Processing}, pages 7798--7809. Association for Computational
  Linguistics.

\bibitem[{Thorne et~al.(2018)Thorne, Vlachos, Christodoulopoulos, and
  Mittal}]{thorne:fever}
James Thorne, Andreas Vlachos, Christos Christodoulopoulos, and Arpit Mittal.
  2018.
\newblock \href {https://doi.org/10.18653/v1/n18-1074} {Fever: a large-scale
  dataset for fact extraction and verification}.
\newblock In \emph{Proceedings of the 2018 Conference of the North American
  Chapter of the Association for Computational Linguistics:Human Language
  Technologies}, pages 809--819. Association for Computational Linguistics.

\bibitem[{Tu et~al.(2020)Tu, Huang, Wang, Huang, He, and Zhou}]{tu:graph2}
Ming Tu, Kevin Huang, Guangtao Wang, Jing Huang, Xiaodong He, and Bowen Zhou.
  2020.
\newblock \href {https://ojs.aaai.org/index.php/AAAI/article/view/6441}
  {Select, answer and explain: Interpretable multi-hop reading comprehension
  over multiple documents}.
\newblock In \emph{Proceedings of the 34th AAAI Conference on Artificial
  Intelligence}, pages 9073--9080. AAAI Press.

\bibitem[{Tu et~al.(2019)Tu, Wang, Huang, Tang, He, and Zhou}]{tu:graph1}
Ming Tu, Guangtao Wang, Jing Huang, Yun Tang, Xiaodong He, and Bowen Zhou.
  2019.
\newblock \href {https://doi.org/10.18653/v1/p19-1260} {Multi-hop reading
  comprehension across multiple documents by reasoning over heterogeneous
  graphs}.
\newblock In \emph{Proceedings of the 57th Conference of the Association for
  Computational Linguistics}, pages 2704--2713. Association for Computational
  Linguistics.

\bibitem[{Vo and Lee(2019)}]{Vo:user-role}
Nguyen Vo and Kyumin Lee. 2019.
\newblock \href {https://doi.org/10.1145/3331184.3331248} {Learning from
  fact-checkers: Analysis and generation of fact-checkinglanguage}.
\newblock In \emph{Proceedings of the 42nd International ACM SIGIR Conference
  on Research and Development in Information Retrieval}, pages 335--344. ACM.

\bibitem[{Ye et~al.(2020)Ye, Lin, Du, Liu, Li, Sun, and Liu}]{ye:corefbert}
Deming Ye, Yankai Lin, Jiaju Du, Zhenghao Liu, Peng Li, Maosong Sun, and
  Zhiyuan Liu. 2020.
\newblock \href {https://doi.org/10.18653/v1/2020.emnlp-main.582}
  {Coreferential reasoning learning for language representation}.
\newblock In \emph{Proceedings of the 2020 Conference on Empirical Methods in
  Natural Language Processing}, pages 7170--7186. Association for Computational
  Linguistics.

\bibitem[{Zhao et~al.(2020)Zhao, Xiong, Rosset, Song, Bennett, and
  Tiwary}]{zhao:xh}
Chen Zhao, Chenyan Xiong, Corby Rosset, Xia Song, Paul~N. Bennett, and Saurabh
  Tiwary. 2020.
\newblock \href {https://openreview.net/forum?id=r1eIiCNYwS} {Transformer-xh:
  Multi-evidence reasoning with extra hop attention}.
\newblock In \emph{Proceedings of the 8th International Conference on Learning
  Representations}. OpenReview.net.

\bibitem[{Zhong et~al.(2020)Zhong, Xu, Tang, Xu, Duan, Zhou, Wang, and
  Yin}]{zhong:dream}
Wanjun Zhong, Jingjing Xu, Duyu Tang, Zenan Xu, Nan Duan, Ming Zhou, Jiahai
  Wang, and Jian Yin. 2020.
\newblock \href {https://doi.org/10.18653/v1/2020.acl-main.549} {Reasoning over
  semantic-level graph for fact checking}.
\newblock In \emph{Proceedings of the 58th Annual Meeting of the Association
  for Computational Linguistics}, pages 6170--6180. Association for
  Computational Linguistics.

\bibitem[{Zhou et~al.(2019)Zhou, Han, Yang, Liu, Wang, Li, and Sun}]{zhou:gear}
Jie Zhou, Xu~Han, Cheng Yang, Zhiyuan Liu, Lifeng Wang, Changcheng Li, and
  Maosong Sun. 2019.
\newblock \href {https://doi.org/10.18653/v1/p19-1085} {Gear: Graph-based
  evidence aggregating and reasoning for fact verification}.
\newblock In \emph{Proceedings of the 57th Conference of the Association for
  Computational Linguistics}, pages 892--901. Association for Computational
  Linguistics.

\bibitem[{Zlatkova et~al.(2019)Zlatkova, Nakov, and
  Koychev}]{Zlatkova:multimodal}
Dimitrina Zlatkova, Preslav Nakov, and Ivan Koychev. 2019.
\newblock \href {https://doi.org/10.18653/v1/D19-1216} {Fact-checking meets
  fauxtography: Verifying claims about images}.
\newblock In \emph{Proceedings of the 2019 Conference on Empirical Methods in
  Natural Language Processing and the 9th International Joint Conference on
  Natural Language Processing}, pages 2099--2108. Association for Computational
  Linguistics.

\end{thebibliography}

\end{document}